\documentclass[11pt,a4paper]{article}
\usepackage[nohyperref]{naaclhlt2018}
\usepackage{times}
\usepackage{latexsym}
\usepackage[utf8]{inputenc}
\usepackage{multirow}

\usepackage{url}
\usepackage{subcaption}
\usepackage{tikz}
\usetikzlibrary{automata,positioning}
\tikzset{initial text={}}

\DeclareCaptionFont{10pt}{\fontsize{10pt}{12pt}\selectfont}
\aclfinalcopy 
\def\aclpaperid{778} 

\title{Neural Machine Translation Decoding with Terminology Constraints}

\author{Eva Hasler$^{1}$, Adrià de Gispert$^{1,2}$, Gonzalo Iglesias$^{1}$, Bill Byrne$^{1,2}$ \\\\
    $^1$SDL Research, Cambridge, UK \\
    $^2$Department of Engineering, University of Cambridge, UK \\
    {\tt \{ehasler,agispert,giglesias,bbyrne\}@sdl.com} \\
}
  
\date{}

\begin{document}
\maketitle
\begin{abstract}
  Despite the impressive quality improvements yielded by neural machine translation (NMT) systems, controlling their translation output to adhere to user-provided terminology constraints remains an open problem. We describe our approach to constrained neural decoding based on finite-state machines and multi-stack decoding which supports target-side constraints as well as constraints with corresponding aligned input text spans. We demonstrate the performance of our framework on multiple translation tasks and motivate the need for constrained decoding with attentions as a means of reducing misplacement and duplication when translating user constraints.
\end{abstract}

\section{Introduction}
Adapting an NMT system with domain-specific data is one way to adjust its output vocabulary to better match the target domain \cite{FineTuning,Backtranslation}. Another way to encourage the beam decoder to produce certain words in the output is to explicitly reward n-grams provided by an SMT system \cite{LNMT} or language model \cite{NmtWithLms} or to modify the vocabulary distribution of the decoder with suggestions from a terminology \cite{GuidedNeuralMT}. While providing lexical guidance to the decoder, these methods do not strictly enforce a terminology. This is a requisite, however, for companies wanting to ensure that brand-related information is rendered correctly and consistently when translating web content or manuals and is often more important than translation quality alone. Although domain adaptation and guided decoding can help to reduce errors in these use cases, they do not provide reliable solutions.

Another recent line of work strictly enforces a given set of words in the output \cite{GuidedImageCaptioning,ConstrainedDecoding,Systran}. \citeauthor{GuidedImageCaptioning} address the task of image captioning with \emph{constrained beam search} where constraints are given by image tags and constraint permutations are encoded in a finite-state acceptor (FSA).
\citeauthor{ConstrainedDecoding} propose \emph{grid beam search} to enforce target-side constraints for domain adaptation via terminology. 
However, since there is no correspondence between constraints and the source words they cover, correct constraint placement is not guaranteed and the corresponding source words may be translated more than once. \citeauthor{Systran} replace entities with special tags that remain unchanged during translation and are replaced in a post-processing step using attention weights. Given good alignments, this method can translate entities correctly but it requires training data with entity tags and excludes the entities from model scoring. 

We address decoding with constraints to produce translations that respect the terminologies  of corporate customers while maintaining the high quality of unconstrained translations. To this end, we apply the constrained beam search of \citeauthor{GuidedImageCaptioning} to machine translation and propose to employ alignment information between target-side constraints and their corresponding source words. The lack of explicit alignments in NMT systems poses an extra challenge compared to statistical MT where alignments are given by translation rules. We address the problem of \emph{constraint placement} by expanding constraints when the NMT model is attending to the correct source span. We also reduce \emph{output duplication} by masking covered constraints in the NMT attention model.

\section{Constrained Beam Search}
A naive approach to decoding with constraints would be to use a large beam size and select from the set of complete hypotheses the best that satisfies all constraints. However, this is infeasible in practice because it would require searching a potentially very large space to ensure that even hypotheses with low model score due to the inclusion of a constraint would be part of the set of outputs. A better strategy is to force the decoder to produce hypotheses that satisfy the constraints regardless of their score and thus guide the decoder into the right area of the search space. We follow \citet{GuidedImageCaptioning} in organizing our beam search into multiple stacks corresponding to subsets of satisfied constraints as defined by FSA states.

\subsection{Finite-state Acceptors for Constraints}
Before decoding, we build an FSA defining the constrained target language for an input sentence. It contains all permutations of constraints interleaved with loops over the remaining vocabulary. 

{\bf Phrase Constraints:} Constraints consisting of multiple tokens are encoded by one state per token. We refer to states within a phrase as intermediate states and restrict their outgoing vocabulary to the next token in the phrase.

{\bf Alternative Constraints:} Synonyms of constraints can be defined as alternatives and encoded as different arcs connecting the same states. When alternatives consist of multiple tokens, the alternative paths will contain intermediate states.

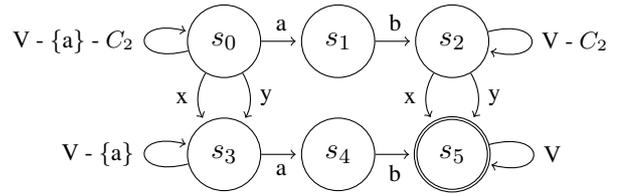
\begin{figure}
\hspace*{-3mm}
\begin{tikzpicture}[shorten >=2pt,node distance=1.5cm,on grid,auto] 
   \node[state] (s_0)   {$s_0$};
   \node[state] (s_11) [right=of s_0] {$s_1$}; 
   \node[state] (s_1) [right=of s_11] {$s_2$};
   \node[state] (s_2) [below=of s_0] {$s_3$};
   \node[state](s_33) [right=of s_2] {$s_4$};
   \node[state,accepting](s_3) [below=of s_1] {$s_5$};
     \path[->] 
     (s_0) edge  node {\small a} (s_11)
           edge [bend right, swap] node {\small x} (s_2)
           edge [bend left] node {\small y} (s_2)
           edge [loop left] node {\small V - \{a\} - $C_2$} ()
     (s_11) edge  node {\small b} (s_1)
     (s_1) edge [bend right, swap] node  {\small x} (s_3)
		   edge [bend left] node  {\small y} (s_3)
           edge [loop right] node {\small V - $C_2$} ()
     (s_2) edge  node [swap] {\small a} (s_33)
           edge [loop left] node {\small V - \{a\}} ()
     (s_33) edge [swap] node {\small b} (s_3)
     (s_3) edge [loop right] node {\small V} ();
\end{tikzpicture}
\caption{\label{fig:acceptor} Example of FSA for two constraints $C_1=a b$ and $C_2=\{x,y\}$.}
\end{figure}

Figure~\ref{fig:acceptor} shows an FSA with constraints $C_1$ and $C_2$ where $C_1$ is a phrase (yielding intermediate states $s_1$, $s_4$) and $C_2$ consists of two single-token alternatives. Both permutations $C_1 C_2$ and $C_2 C_1$ lead to final state $s_5$ with both constraints satisfied.

\subsection{Multi-Stack Decoding}
When extending a hypothesis to satisfy a constraint which is not among the top-$k$ vocabulary items in the current beam, the overall likelihood may drop and the hypothesis may be pruned in subsequent steps. To prevent this, the extended hypothesis is placed on a new stack along with other hypotheses that satisfy the same set of constraints. Each stack maps to an acceptor state which helps to keep track of the permitted extensions for hypotheses on this stack. The stack where a hypothesis should be placed is found by following the appropriate arc leaving the current acceptor state. The stack mapping to the final state is used to generate complete hypotheses. At each time step, all stacks are pruned to the beam size $k$ and therefore the actual beam size for constrained decoding depends on the number of acceptor states. 

\subsection{Decoding with Attentions}
\label{sec:approach}
Since an acceptor encoding $c$ single-token constraints has $2^c$ states, the constrained search of \citet{GuidedImageCaptioning} can be inefficient for large numbers of constraints. In particular, all unsatisfied constraints are expanded at each time step $t$ which increases decoding complexity from $\mathcal{O}(tk)$ for normal beam search to $\mathcal{O}(tk2^c)$. \citet{ConstrainedDecoding} organize their grid beam search into beams that group hypotheses with the same number of constraints, thus their decoding time is $\mathcal{O}(tkc)$. However, this means that different constraints will compete for completion of the same hypothesis and their placement is determined locally. We assume that a target-side constraint can come with an aligned source phrase which is encoded as a span in source sentence $S$ and stored with the acceptor arc label:

\begin{center}
\begin{small}
\begin{tikzpicture}[shorten >=1pt,node distance=2cm,on grid,auto] 
   \node[state] (s_0)   {$s_0$}; 
   \node[state] (s_1) [right=of s_0, label={right: $\quad j > i, \quad 0 \leq i,j \leq |S|$}] {$s_1$};
     \path[->] 
     (s_0) edge  node {\small $C$ [i,j)} (s_1);
\end{tikzpicture}
\end{small}
\end{center}

Because the attention weights in attention-based decoders function as soft alignments from the target to the source sentence \cite{Alkouli2017}, we use them to decide at which position a constraint should be inserted in the output. At each time step in a hypothesis, we determine the source position with the maximum attention. If it falls into a constrained source span and this span matches an outgoing arc in the current acceptor state, we extend the current hypothesis with the arc label. Thus, the outgoing arcs in non-intermediate states are active or inactive depending on the current attentions. This reduces the complexity from $\mathcal{O}(tk2^c)$ to $\mathcal{O}(tkc)$ by ignoring all but one constraint permutation and in practice, disabling vocabulary loops saves extra time.

{\bf State-specific Attention Mechanism:}
Once a constraint has been completed, we need to ensure that its source span will not be translated again. We force the decoder to respect covered constraints by masking their spans during all future expansions of the hypothesis. This is done by zeroing out the attention weights on covered positions to exclude them from the context vector computed by the attention mechanism.

{\bf Implications:} Constrained decoding with aligned source phrases relies on the quality of the source-target pairs. Over- and under-translation can occur as a result of incomplete source or target phrases in the terminology.

{\bf Special Cases:} Monitoring the source position with the maximum attention is a relatively strict criterion to decide where a constraint should be placed in the output. It turns out that depending on the language pair, the decoder may produce translations of neighbouring source tokens when attending to a constrained source span.\footnote{For example, to produce an article before a noun when the constrained source span includes just the noun.} The strict requirement of only producing constraint tokens can be relaxed to accommodate such cases, for example by allowing extra tokens before ($s_1$) or after ($s_2$) constraint $C$ while attending to span $[i,j)$, 
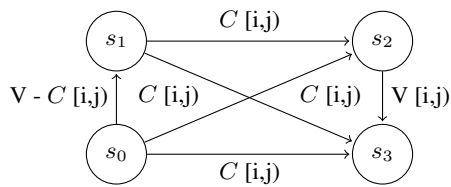
\begin{center}
\begin{small}
\begin{tikzpicture}[shorten >=1pt,node distance=1.5cm,on grid,auto] 
   \node[state] (s_0) {$s_0$};
   \node[state] (s_1) [above=of s_0] {$s_1$}; 
   \node[state] (s_2) [right=3.5cm of s_1] {$s_2$};
   \node[state] (s_3) [right=3.5cm of s_0] {$s_3$};
     \path[->] 
     (s_0) edge  node {\small V - $C$ [i,j)} (s_1)
     (s_0) edge  node [pos=0.3] {\small $C$ [i,j)} (s_2) 
     (s_0) edge [swap] node {\small $C$ [i,j)} (s_3)
     (s_1) edge  node {\small $C$ [i,j)} (s_2)
     (s_1) edge  node [pos=0.7] {\small $C$ [i,j)} (s_3) 
     (s_2) edge  node {\small V [i,j)} (s_3);
\end{tikzpicture}
\end{small}
\end{center}
Conversely, the decoder may never place the maximum attention on a constraint span which can lead to empty translations. Relaxing this requirement using thresholding on the attention weights to determine positions with secondary attention can help in those cases.

\section{Experimental Setup}
\label{experiments}
We build attention-based neural machine translation models \cite{Bahdanau2015} using the Blocks implementation of \citet{Blocks} for English-German and English-Chinese translation in both directions. We combine three models per language pair as ensembles and further combine the NMT systems with n-grams extracted from SMT lattices using Lattice minimum Bayes-risk as described by \citet{LNMT}, referred to as \textsc{Lnmt}. We decode with a beam size of 12 and length normalization \cite{GNMT} and back off to constrained decoding without attentions when decoding with attentions fails.\footnote{This usually applies to less than 2\% of the inputs.} We report lowercase \textsc{Bleu} using mteval-v13.pl.

\subsection{Data}
Our models are trained on the data provided for the 2017 Workshop for Machine Translation \cite{WMT17}. We tokenize and truecase the English-German data and apply compound splitting when the source language is German. The training data for the NMT systems is augmented with backtranslation data \cite{Backtranslation}. For English-Chinese, we tokenize and lowercase the data. We apply byte-pair encoding \cite{BPE} to all data.

\subsection{Terminology Constraints}
We run experiments with two types of constraints to evaluate our constrained decoder.

{\bf Gold Constraints:} For each input sentence, we extract up to two tokens from the reference which were not produced by the baseline system, favouring rarer words. This aims at testing the performance in a setup where users may provide corrections to the NMT output which are to be incorporated into the translation. These reference tokens may consist of one or more subwords. Similarly, we extract phrases of up to five subwords surrounding a reference token missing from the baseline output. We do not have access to aligned source words for gold constraints.

{\bf Dictionary Constraints:} 
We automatically extract bilingual dictionary entries using terms and phrases from the reference translations as candidates in order to ensure that the entries are relevant for the inputs. In a real setup, the dictionaries would be provided by customers and would be expected to contain correct translations without ambiguity. We apply a filter of English stop words and verbs to the candidates and look them up in a pruned phrase table to find likely pairs, resulting in entries as shown below:\footnote{Our dictionaries are available on request.}\\

\begin{small}
\begin{tabular}{ll}
\bf English & \bf German \\
ICJ & IGH \\
The Wall Street Journal & The Wall Street Journal \\
Dead Sea & Tote Meer$|$Toten Meer \\\\
\end{tabular}
\end{small}

For evaluation purposes, we ensure that dictionary entries match the reference when applying them to an input sentence.

\setlength{\tabcolsep}{3pt}
\begin{table*}[t!]
\begin{subtable}{.6\linewidth}\centering
{\begin{tabular}{lc|c|ccc}
& \bf dev (lr) & \bf rep & \bf test15 & \bf test16 & \bf test17 \\\hline
\multicolumn{1}{l}{\em eng-ger-wmt17} \\\hline
\textsc{Lnmt} & 24.9 (1.00) & 443 & 28.1 & 34.7 & 27.0 \\
+ 2 gold tokens & 29.2 (1.14) & 1141 & 33.4 & 40.9 & 32.3 \\
+ 1 gold phrase & 36.8 (1.09) & 880 & 40.5 & 46.7 & 39.6 \\ 
\rule{0pt}{3ex}+ dictionary (v1) & 26.4 (1.03) & 610 & 29.6 & 36.4 & 28.8 \\ 
+ dictionary (v2) & 26.6 (1.02) & 471 & 29.9 & 37.0 & 29.1 \\\hline
\multicolumn{1}{l}{\em ger-eng-wmt17} \\\hline
\textsc{Lnmt} & 31.2 (1.01) & 307 & 33.5 & 40.7 & 34.6 \\
+ 2 gold tokens & 34.6 (1.14) & 745 & 37.7 & 44.8 & 38.5 \\
+ 1 gold phrase & 42.3 (1.08) & 550 & 45.7 & 51.3 & 46.4 \\
\rule{0pt}{3ex}+ dictionary (v1) & 32.4 (1.02) & 353 & 34.7 & 41.8 & 36.2 \\ 
+ dictionary (v2) & 32.5 (1.01) & 320 & 34.6 & 41.9 & 36.0 \\\hline
\end{tabular}}
\caption{Results for English-German language pairs}\label{tab:1a}
\end{subtable}%
\begin{subtable}{.4\linewidth}\centering
{\begin{tabular}{l|c|c}
& \bf dev (lr) & \bf test17 \\\hline
\multicolumn{1}{l}{\em eng-chi-wmt17} \\\hline
\textsc{Lnmt} & 30.8 (0.95) & 31.0 \\
+ 2 gold tokens & 33.8 (1.10) & 34.2 \\
+ 1 gold phrase & 40.6 (1.06) & 41.2 \\
\rule{0pt}{3ex}+ dictionary (v1) & 34.0 (1.01) & 33.7 \\ 
+ dictionary (v2) & 33.9 (0.98) & 34.1 \\\hline
\multicolumn{1}{l}{\em chi-eng-wmt17} \\\hline
\textsc{Lnmt} & 21.2 (1.00) & 23.5 \\
+ 2 gold tokens & 23.3 (1.13) & 25.5 \\
+ 1 gold phrase & 30.1 (1.09) & 32.3 \\
\rule{0pt}{3ex}+ dictionary (v1) & 23.0 (1.06) & 25.5 \\ 
+ dictionary (v2) & 23.4 (1.03) & 25.4 \\\hline
\end{tabular}}
\caption{Results for English-Chinese language pairs}\label{tab:1b}
\end{subtable}
\caption{\textsc{Bleu} scores and dev length ratios for decoding with gold constraints (without attentions) followed by results for dictionary constraints without (v1) or with (v2) attentions. The column \emph{rep} shows the number of character 7-grams that occur more than once within a sentence of the dev set, see Section~\ref{sec:repetitions}.}\label{tab:1}
\end{table*}

\section{Results}
The results for decoding with terminology constraints are shown in Table~\ref{tab:1a} and~\ref{tab:1b} where each section contains the results for gold constraints followed by dictionary constraints.

\subsection{Results with Gold Constraints}
\begin{table*}[h!]
\centering
\small
\begin{tabular}{ p{2.4cm} | p{5.5cm} | p{7cm}  }
\hline\multicolumn{1}{l}{\em eng-ger-wmt17} & \em Example 1 &\em Example 2 \\\hline
Source & It already has the {\bf budget} ... & And it often costs over a hundred dollars to obtain the required {\bf identity card}.\\
\rule{0pt}{2ex}Constraints & Budget [4,5) & Ausweis [12,14)\\
\rule{0pt}{3ex}\textsc{Lnmt} & Es hat bereits den {\bf Haushalt}... & Und es kostet oft mehr als hundert Dollar, um die erforderliche {\bf Personalausweis} zu erhalten.\\
\rule{0pt}{2ex}+ dictionary (v1) & Das {\bf Budget} hat bereits den {\bf Haushalt}... & Und es kostet oft mehr als hundert Dollar, um den {\bf Ausweis} zu erhalten, um die erforderliche {\bf Personalausweis} zu erhalten.\\
\rule{0pt}{2ex}+ dictionary (v2) & Es verfügt bereits über das {\bf Budget}... & Und es kostet oft mehr als hundert Dollar, um den gewünschten {\bf Ausweis} zu erhalten.\\\hline
\multicolumn{1}{l}{\em ger-eng-wmt17} & \em Example 3 & \em Example 4 \\\hline
Source & Der {\bf Pokal} war die einzige {\bf Möglichkeit} , etwas zu gewinnen . & Aber es ist keine typische Zeichensprache -- sagt sie . Edmund hat einige {\bf Zeichen} alleine erfunden .\\
\rule{0pt}{2ex}Constraints & cup [1,2), chance [5,6) & sign$|$signs [13,14) \\
\rule{0pt}{3ex}\textsc{Lnmt} & The {\bf trophy} was the only {\bf way} to win something. & But it's not a typical sign language -- says, Edmund invented some {\bf characters} alone.\\
\rule{0pt}{2ex}+ dictionary (v1) & The {\bf cup} was the only {\bf way} to get something to win a {\bf chance}. & But it's not a typical sign language -- says, Edmund invented some {\bf characters} alone. \\
\rule{0pt}{2ex}+ dictionary (v2) & The {\bf cup} was the only {\bf chance} to win something. & But it is not a typical sign language -- she says, Edmund invented some {\bf signs} alone. \\
\end{tabular}
\caption{English$\leftrightarrow$German translation outputs for constrained decoding.}\label{tab:examples}
\end{table*}

Decoding with gold constraints yields large \textsc{Bleu} gains over \textsc{Lnmt} for all language pairs. However, the length ratio on the dev set increases significantly. Inspecting the output reveals that this is often caused by constraints being translated more than once which can lead to whole passages being retranslated. Phrase constraints seem to integrate better into the output than single token constraints which may be due to the longer gold context being fed back to the \textsc{Nmt} state.

\subsection{Results with Dictionary Constraints}
Decoding with up to two dictionary constraints per sentence yields gains of up to 3 \textsc{Bleu}. This is partly because we do not control whether \textsc{Lnmt} already produced the constraint tokens and because not all sentences have dictionary matches. The length ratios are better compared to the gold experiments which we attribute to our filtering of tokens such as verbs which tend to influence the general word order more than nouns, for example. 

Decoding with or without attentions yields similar \textsc{Bleu} scores overall and a consistent improvement for English-German. Note that decoding with attentions is sensitive to errors in the automatically extracted dictionary entries. 

{\bf Output Duplication} The first three examples in Table~\ref{tab:examples} show English$\leftrightarrow$German translations where decoding without attentions has generated both the target side of the constraint and the translation preferred by the NMT system. When using the attentions, each constraint is only translated once.

{\bf Constraint Placement} The fourth example demonstrates the importance of tying constraints to source words. Decoding without attentions fails to translate \emph{Zeichen} as \emph{signs} because the alternative \emph{sign} already appears in the translation of \emph{Zeichensprache} as \emph{sign language}. When using the attentions, \emph{signs} is generated at the correct position in the output. 

\subsection{Output length ratio and repetitions}
\label{sec:repetitions}
To back up our hypothesis that increases in length ratio are related to output duplication, Table~\ref{tab:1a} column \emph{rep} shows the number of repeated character 7-grams within a sentence of the dev set, ignoring stop words and overlapping n-grams. This confirms that constrained decoding with attentions reduces the number of repeated n-grams in the output. While this does not account for alignments to the source or capture duplicated translations with unrelated surface forms, it provides evidence that the outputs are not just shorter than for decoding without attentions but in fact contain fewer repetitions and likely fewer duplicated translations.

\newcommand*{\MyIndent}{\hspace*{0.2cm}}%
\begin{table}[h!]
\centering
\begin{tabular}{l|cc|cc|cc}
& \multicolumn{6}{c}{\bf \textsc{Bleu}/speed ratio}  \\\hline
\multicolumn{1}{l}{\em eng-ger-wmt17} & \multicolumn{2}{c}{$c$=2} & \multicolumn{2}{c}{$c$=3} & \multicolumn{2}{c}{$c$=4} \\\hline
\textsc{Lnmt} & 26.7 & 1.00 & 26.7 & 1.00 & 26.7 & 1.00 \\
+ dict (v1) & 28.2 & 0.20 & 28.4 & 0.14 & 28.5 & 0.11 \\ 
+ dict (v2$^{*}$) & 27.8 & 0.69 & 28.0 & 0.66 & 28.1 & 0.59 \\
\MyIndent + A & 28.0 & 0.65 & 28.2 & 0.61 & 28.2 & 0.54 \\
\MyIndent\MyIndent + B & 28.4 & 0.27 & 28.6 & 0.24 & 28.7 & 0.21 \\
\MyIndent\MyIndent + C & 28.5 & 0.21 & 28.6 & 0.19 & 28.7 & 0.17 \\
\end{tabular}
\caption{\textsc{Bleu} scores and speed ratios relative to unconstrained \textsc{Lnmt} for production system with up to $c$ constraints per sentence (newstest2017). A: secondary attention, B, C: allow 1 or 2 extra tokens, respectively (Section~\ref{sec:approach}). \emph{Dict (v2$^{*}$)} refers to decoding with attentions but without A, B or C.}\label{tab:2}
\end{table}

\subsection{Comparison of decoding speeds}
To evaluate the speed of constrained decoding with and without attentions, we decode newstest- 2017 on a single GPU using our English-German production system \cite{LNMT2018} which in comparison to the systems described in Section~\ref{experiments} uses a beam size of 4 and an early pruning strategy similar to that described in \citet{GNMT}, amongst other differences. About 89\% of the sentences have at least one dictionary match and we allow up to two, three or four matches per sentence. Because the constraints result from dictionary application, the number of constraints per sentence varies and not all sentences contain the maximum number of constraints.

Tab.~\ref{tab:2} reports \textsc{Bleu} and speed ratios for different decoding configurations.
Rows two and three confirm that the reduced computational complexity of our approach yields faster decoding speeds than the approach of \citet{GuidedImageCaptioning} while incurring a small decrease in \textsc{Bleu}. Moreover, it compares favourably for larger numbers of constraints per sentence: v2* is 3.5x faster than v1 for $c$=2 and more than 5x faster for $c$=4. Relaxing the restrictions of decoding with attentions improves the \textsc{Bleu} scores but increases runtime. However, the slowest v2 configuration is still faster than v1. The optimal trade-off between quality and speed is likely to differ for each language pair.

\section{Conclusion}
We have presented our approach to NMT decoding with terminology constraints using decoder attentions which enables reduced output duplication and better constraint placement compared to existing methods. Our results on four language pairs demonstrate that terminology constraints as provided by customers can be respected during NMT decoding while maintaining the overall translation quality. 
At the same time, empirical results confirm that our improvements in computational complexity translate into faster decoding speeds.
Future work includes the application of our approach to more recent architectures such as \citet{AIAYN} which will involve extracting attentions from multiple decoding layers and attention heads.

\section*{Acknowledgments}
This work was partially supported by U.K. EPSRC grant EP/L027623/1.

\bibliography{naaclhlt2018}
\bibliographystyle{acl_natbib}




\end{document}